\documentclass{article}

\usepackage{amsmath}
\usepackage{graphicx}
\usepackage[authoryear]{natbib}
\usepackage{ifpdf}
\ifpdf
    \usepackage{epstopdf}%
\fi
\usepackage{multirow}
\usepackage{changepage}
\usepackage{hyperref}
\usepackage{cleveref}
\usepackage{booktabs}
\usepackage[table]{xcolor}
\usepackage{subcaption}
\usepackage{enumitem}
\usepackage{makecell}
\usepackage{authblk}

\title{Evaluating Metalinguistic Knowledge in Large Language Models across the World's Languages}

\author[1]{Tjaša Arčon}
\author[1]{Matej Klemen}
\author[1]{Marko Robnik-Šikonja}
\author[1,2,3]{Kaja Dobrovoljc}

\affil[1]{University of Ljubljana, Faculty of Computer and Information Science, Ljubljana, Slovenia}
\affil[2]{University of Ljubljana, Faculty of Arts, Ljubljana, Slovenia}
\affil[3]{Jožef Stefan Institute, Ljubljana, Slovenia}

\affil[*]{\textbf{Corresponding author:} Tjaša Arčon\\
Email: \href{mailto:tjasa.arcon@fri.uni-lj.si}{tjasa.arcon@fri.uni-lj.si}}

\begin{document}
\label{firstpage}
\maketitle
\vspace{-1.5em}

\begin{abstract}
 
Large language models (LLMs) are routinely evaluated on language use tasks, yet their explicit knowledge about linguistic structure remains poorly understood. Existing linguistic benchmarks typically focus on narrow phenomena, emphasize high-resource languages, and rarely evaluate metalinguistic knowledge—explicit reasoning about language structure rather than language use. In this paper, we present a comprehensive multilingual evaluation of metalinguistic knowledge in LLMs, based on the World Atlas of Language Structures (WALS), a large database of 192 linguistic features across the world's 2,660 languages. We convert WALS features into natural-language questions with predefined answer options and evaluate model performance across the full set of documented languages.
Using accuracy and macro $F_1$, together with majority-class and chance baselines, we analyse overall performance and examine variation by linguistic domains and language-related factors. 
Our results show that metalinguistic knowledge in current LLMs is limited: GPT-4o performs best but still achieves only moderate accuracy (0.367), while open-source models lag behind. All models perform above chance but fail to outperform the majority-class baseline, suggesting they capture broad cross-linguistic patterns but lack fine-grained grammatical distinctions. Performance varies across linguistic domains, with lexical features showing the highest accuracy and phonological features among the lowest, partially reflecting differences in online visibility. At the language level, accuracy shows a strong and consistent association with digital language status: languages with higher digital presence and resource availability are evaluated more accurately, while low-resource languages exhibit substantially lower performance. Analyses of predictive factors confirm that resource-related indicators (Wikipedia size, corpus availability) are more informative predictors of model accuracy than geographical, genealogical or sociolinguistic factors. Together, these results suggest that LLMs’ metalinguistic knowledge is fragmented and strongly shaped by data availability, rather than reflecting broadly generalizable grammatical competence across the world’s languages. We release our benchmark as an open-source dataset to support systematic evaluation of metalinguistic knowledge across the world’s languages and to encourage greater global linguistic diversity in future LLMs.
 
\end{abstract}


\section{Introduction}

Large language models (LLMs) are routinely evaluated on tasks ranging from text generation to question answering, but rarely on their explicit knowledge of language structure. In other words, while we know that LLMs can \textit{use} language fluently \citep{Chang2024SurveyLLM}, we know far less about what they know about language itself—a gap that is especially pronounced for low-resource languages, where limited training data may result in even more fragmented or unreliable linguistic representations. Explicit linguistic knowledge includes awareness of grammatical properties such as word order, agreement, case marking, or phonological patterns, which underpin linguistic analysis and explanation. Understanding whether LLMs possess such knowledge is crucial, particularly as they are increasingly employed in linguistically informed tasks such as annotation, grammatical analysis, and cross-linguistic comparison \citep{begus2025large, kellert2025parsing, ramji2025inductive, waldis2024holmes}, as well as in language documentation, where they are used to accelerate transcription, translation, morphological analysis, glossing, and grammatical description, crucial for the preservation of endangered languages \citep{annurev:/content/journals/10.1146/annurev-linguistics-031220-120504,spencer-kongborrirak-2025-llms,tanzer2024benchmarklearningtranslatenew}.

To support these goals, recent research has begun to probe LLMs’ linguistic knowledge through grammatical classification tasks \citep{ide2025make}, feature-specific evaluations and analyses \citep{begus2025large}, as well as an increasing number of targeted benchmarks testing phenomena such as agreement, acceptability, and metalinguistic reasoning \citep{jumelet2025multiblimp,zhang2024mela,behzad2023elqa}. While these efforts provide valuable insights into particular aspects of linguistic competence, they remain narrowly scoped, typically focusing on specific phenomena, tasks, or small language subsets, with a strong emphasis on English and other high-resource languages. As a result, current evaluations provide only a fragmented picture of LLMs' linguistic knowledge and offer little insight into how such knowledge generalizes across the world's languages. This limitation is particularly problematic for low-resource and underdocumented languages, where the lack of systematic evaluation obscures model weaknesses and risks reinforcing existing biases, making models appear linguistically competent while relying primarily on patterns learned from a small number of digitally dominant languages.

\begin{figure}[htbp]
\centering
\includegraphics[width=\linewidth]{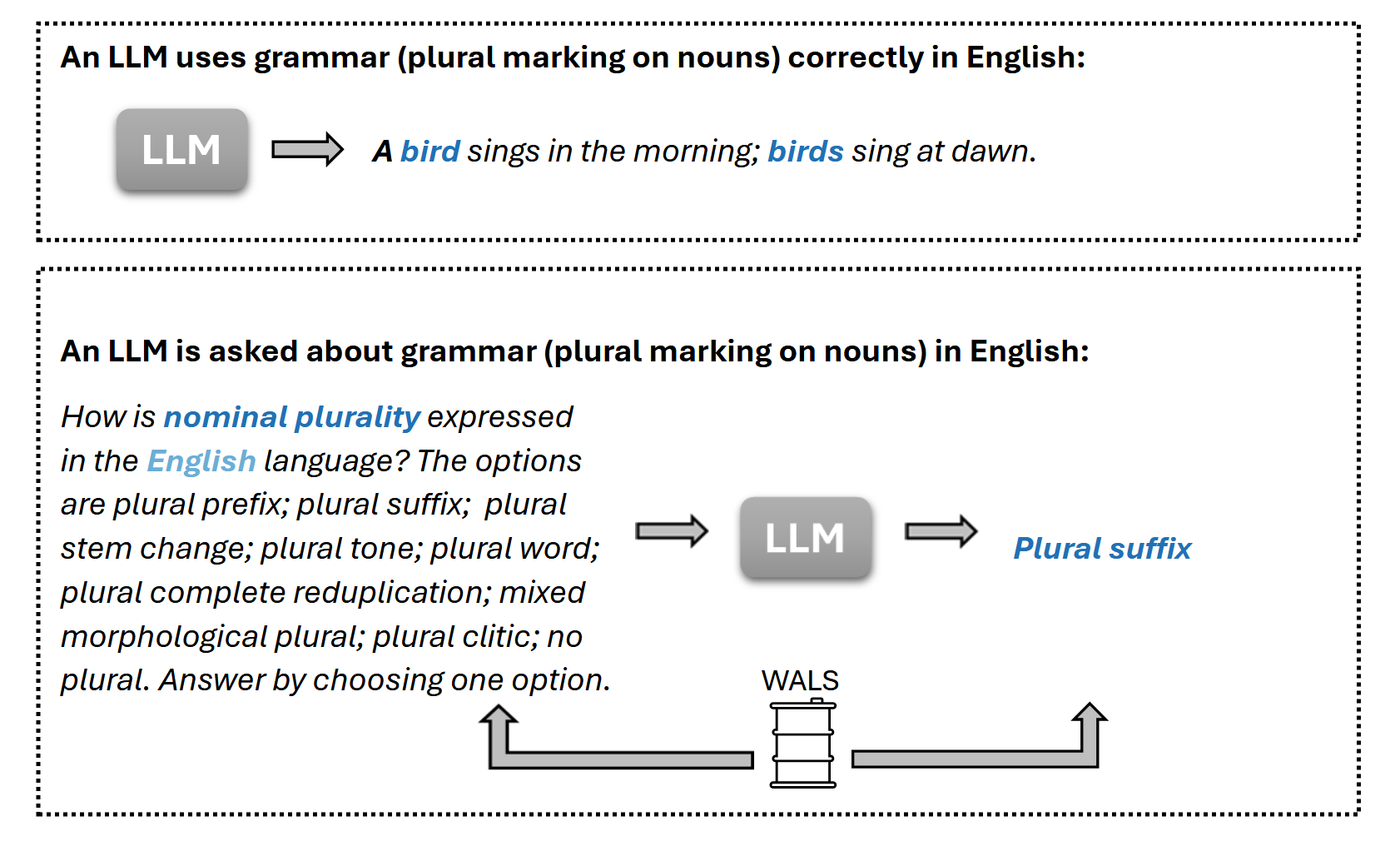}
\caption{High-level overview of our evaluation setup. While LLMs can use grammatical patterns correctly in language generation (top example), we assess their explicit linguistic knowledge by querying models with WALS-based multiple-choice questions and comparing their responses to the corresponding ground-truth feature values documented in WALS (bottom example).}
\label{fig:placeholder}
\end{figure}

To address this gap, we explore the methodological potential of the World Atlas of Language Structures (WALS) \citep{WALS} as a framework for multilingual evaluation of explicit linguistic (metalinguistic) knowledge in LLMs (as illustrated in Figure \ref{fig:placeholder}). WALS documents nearly two hundred grammatical features across more than 2,600 languages, spanning linguistic domains from phonology and morphology to lexicon and syntax, and thus provides a unique basis for large-scale, cross-linguistic evaluation. We systematically convert WALS features into natural-language questions to construct a QA-style benchmark covering all available languages, and use this benchmark to conduct a multidimensional evaluation of several LLMs.

In doing so, we address the following research questions:
\begin{enumerate}[label=\textbf{RQ\arabic*:}, leftmargin=*, itemsep=0.3em]
  \item How accurately do LLMs answer metalinguistic questions about linguistic features across a large and diverse set of languages?
  \item How does LLM performance vary across different linguistic domains?
  \item How does LLM performance vary across languages, and which factors are associated with this variation?
\end{enumerate}

Our results show that metalinguistic knowledge in current LLMs is limited: even the best-performing model achieves only moderate accuracy, while open-source models lag further behind. Performance varies across linguistic domains, with lexical features showing the highest accuracy and phonological features among the lowest, and across languages, with low-resource languages exhibiting substantially lower performance than digitally well-supported ones. These findings highlight the importance of broad, cross-linguistic evaluation when assessing LLMs' linguistic competence. To support such evaluation, our main contributions are as follows:

\begin{enumerate}
  \item \textbf{New multilingual benchmark:} We introduce a massively multilingual benchmark for evaluating explicit linguistic (metalinguistic) knowledge in LLMs, grounded in the World Atlas of Language Structures.
  \item \textbf{Large-scale evaluation:} Using this benchmark, we conduct a large-scale evaluation covering 2,660 languages—including a substantial proportion of low-resource and under-documented languages—and analyse how LLM performance varies across domains and across languages with different levels of digital support.
    \item \textbf{Methodological insights:} We discuss limitations of using WALS for metalinguistic benchmarking, such as uneven language coverage and categorical feature design, and their implications for future evaluation frameworks.

\end{enumerate}

In the remainder of this paper, Section \ref{sec:related work} provides an overview of existing benchmarks for evaluating linguistic knowledge in LLMs; Section \ref{sec:wals} introduces the WALS database and describes the construction of our benchmark; Section \ref{sec:setup} details the models and evaluation protocol; Section \ref{sec:results} presents the results, and Section \ref{sec:discussion} discusses their implications and outlines directions for future work.

\section{Background and related work}
\label{sec:related work}

LLMs are becoming increasingly important in the scientific methodology \citep{lu2024ai} across a range of fields, including linguistics \citep{klemen2025towards,spencer-kongborrirak-2025-llms,singh-etal-2023-explaining}. As a result, their evaluation has become increasingly important, as it determines how effectively different models handle specific tasks. In practice, such evaluation is typically carried out through benchmarks that allow for systematic comparison between models. This section examines benchmarks that specifically assess linguistic competence and organizes them according to different dimensions of linguistic knowledge. 

Recent evaluation work on LLMs examines multiple aspects of linguistic knowledge, indicating that LLMs’ linguistic ability is better understood as a set of distinct layers rather than a single unified competence. Some available benchmarks focus on explicit grammatical knowledge, testing whether models apply specific grammatical rules and distinguish between correct and incorrect grammatical usage (Section \ref{sec:rw-grammatical}). Other benchmarks scrutinize metalinguistic competence, assessing the ability of LLMs to act as linguists by reasoning explicitly about language, identifying linguistic structures, or performing linguistic analyses in different languages (Section \ref{sec:rw-metalinguistic}). A third group of benchmarks focuses on pedagogical linguistic knowledge, treating LLMs as potential language teachers that can explain a wide variety of grammatical rules in different languages (Section \ref{sec:rw-pedagogical}). Finally, certain benchmarks, such as Holmes \citep{waldis2024holmes}, investigate the linguistic performance of LLMs at the level of internal embeddings using probing instead of evaluating observable outputs through prompting. As the present work is concerned with linguistic knowledge that is accessible through direct interaction with LLMs, we restrict our survey below to benchmarks based on prompting rather than internal probing techniques. Accordingly, the following overview is organized from benchmarks that assess surface grammatical competence to those targeting metalinguistic and pedagogical evaluations.

\subsection{Evaluation of grammatical competence}
\label{sec:rw-grammatical}
Several benchmarks investigate the surface linguistic capability. \citet{jumelet2025multiblimp} compile a massively multilingual benchmark, MultiBLiMP 1.0, consisting of minimal pairs that test formal grammatical knowledge, evaluating morphosyntactic subject-verb and subject-participle agreement for number, person, and gender across 101 languages. They evaluate 42 language models on grammatical preference using probability-based differences between minimal pairs. Their results show strong performance for high-resource languages, which drops sharply for low-resource languages, even for larger models that consistently outperform smaller ones.  Accuracy correlates strongly with language frequency in Common Crawl, suggesting that grammatical competence is mainly data-driven and may deteriorate during post-training. 

Similarly, the MELA benchmark \citep{zhang2024mela} assesses whether a model can distinguish between grammatical and ungrammatical sentences, encompassing morphology and syntax features such as word order, agreement, and relative clauses. It measures the linguistic acceptability of presented sentences in ten typologically diverse languages. The findings demonstrate that modern LLMs can perform human-like acceptability judgments across multiple languages, but open-source models lag significantly behind closed models. The benchmark does not include any low-resource language. 

PhonologyBench \citep{suvarna2404phonologybench} is an English-only benchmark that tests how well LLMs understand phonology through the grapheme-to-phoneme task, syllable counting, and rhyme judgement. It occupies an intermediate position between surface linguistic competence and explicit metalinguistic reasoning since rhyme judgment reflects surface phonological behaviour, while the other two tasks require implicit phonological analyses. The benchmark is used to evaluate six major LLMs. The study demonstrates that LLM competence remains below human performance, especially for tasks that require abstract phonological reasoning such as syllable counting. Performance varies widely across models and tasks, although LLMs exhibit some phonological awareness despite being trained on texts, indicating that some phonological structure is indirectly learned from orthography.  

LINGGYM \citep{yang2025linggym} is another benchmark that bridges surface grammatical knowledge and metalinguistic reasoning. The benchmark tests whether models can infer a multiple-choice masked word or word-gloss pair in a sentence based on provided linguistic information, so the models need to apply grammatical descriptions to reconstruct linguistic structure. The benchmark is multi-lingual and spans across eighteen low-resource languages, many of them severely underrepresented. Without grammatical cues models perform only slightly above chance, but with structured linguistic information accuracy improves across all models. However, even strong LLMs show poor performance, especially on unseen languages, complex morphological paradigms, and abstract grammatical rules. 

\subsection{Evaluation of metalinguistic competence}
\label{sec:rw-metalinguistic}
As having the structure is not the same as talking about the structure, some benchmarks test how well LLMs answer metalinguistic questions about different languages. The first publicly available corpus of metalinguistic questions and answers was ELQA \citep{behzad2023elqa}, with over 70,000 metalinguistic questions from English learners, collected from two online Stack Exchange forums, covering topics such as grammar, meaning, fluency, and etymology. In contrast to benchmarks evaluating surface grammaticality, ELQA does not test preference or acceptability, but instead assesses whether models can generate accurate and informative linguistic explanations. The results suggest that, although the LLM outputs are fluent, their linguistic validity and correctness are below human performance. Explanations are often partially incorrect or misleading, with models performing better on meaning-related questions than on explicit grammatical analysis. 

A dataset that evaluates how well models deal with metalinguistic self-reference was developed by \citet{thrush2024strange}. The dataset consists of two subtasks: i) generation, where models continue statements with truth-preserving completions, and ii) verification, where they judge the truth of completed statements. To assess whether models can handle metalinguistic language in general, minimally different metalinguistic control tasks without self-reference are included. The study concludes that models struggle with metalinguistic self-reference and perform at or near chance in all domains. Although GPT-4 shows  improvement, it remains well below human performance.

IOLBENCH \citep{goyal2025iolbench} evaluates a different kind of metalinguistic knowledge by focusing on linguistic reasoning based on puzzles that are derived from the International Linguistics Olympiad (ILO). The benchmark tests whether models can infer grammatical systems from linguistic data and comes to the conclusion that current LLMs struggle with linguistic tasks that require explicit rule induction, especially without prior knowledge. Moreover, LingBench++ \citep{lian2025lingbench++} is also derived from ILO problems targeting inductive linguistic reasoning across over 90 low-resource and typologically diverse languages. Additionally, it measures LLM reasoning quality and analyses how reasoning unfolds. The results indicate that even strong LLMs struggle with abstract grammatical rule induction. They perform worst on the phonological rule system and multi-rule grammatical systems, but slightly better on lexical and morphological pattern matching. 

\subsection{Evaluation of pedagogical competence}
\label{sec:rw-pedagogical}
The third group of benchmarks tests pedagogical linguistic knowledge. CPG-EVAL \citep{wang2025cpg} is the first benchmark designed to measure pedagogical grammar competence of LLM in teaching Chinese as a second language. It checks whether models can correctly recognize and discriminate teaching-oriented grammar rules for Chinese. It emerges that LLMs perform strongly on simple grammar recognition, but their performance drops sharply with increasing task complexity. 

Similarly, a part of the CLTE benchmark \citep{xu2025can} addresses linguistic knowledge as part of a broader benchmark that evaluates the pedagogical competence of LLMs functioning as language teachers for Chinese as a second language. This study also confirms that models struggle with pedagogical competence related to linguistic and grammar explanation as performance is below human teacher standards. 

\subsection{Our contribution}

Across the benchmarks reviewed above, a number of recurring observations have been reported, including better performance of larger models compared to smaller ones, an advantage for high-resource languages over low-resource ones, and higher accuracy on surface-level grammatical tasks than on tasks requiring abstract, metalinguistic, or pedagogical reasoning. However, these observations emerge from heterogeneous benchmarks that differ substantially in task design, linguistic scope, and language coverage, making it difficult to assess the extent to which such patterns generalize across languages and types of linguistic knowledge, making it difficult to assess how robust they are across the world’s languages and across different types of linguistic knowledge—particularly for typologically diverse and low-resource languages, which remain largely underrepresented.

In this work, we advance the state of the art by introducing a new large-scale multilingual benchmark for evaluating metalinguistic knowledge in LLMs and using it for a systematic, multi-dimensional analysis. Our framework enables evaluation across a broad range of languages and linguistic domains, supports principled comparisons across language groups, and allows us to examine how performance varies with linguistic domain, language characteristics, and resource-related factors. This provides a more comprehensive and fine-grained view of LLMs’ metalinguistic capabilities than prior evaluations focused on smaller language samples or isolated phenomena.

\section{Benchmark construction from WALS}
\label{sec:wals}

\begin{figure}[htbp]
\centering
\includegraphics[width=\linewidth]{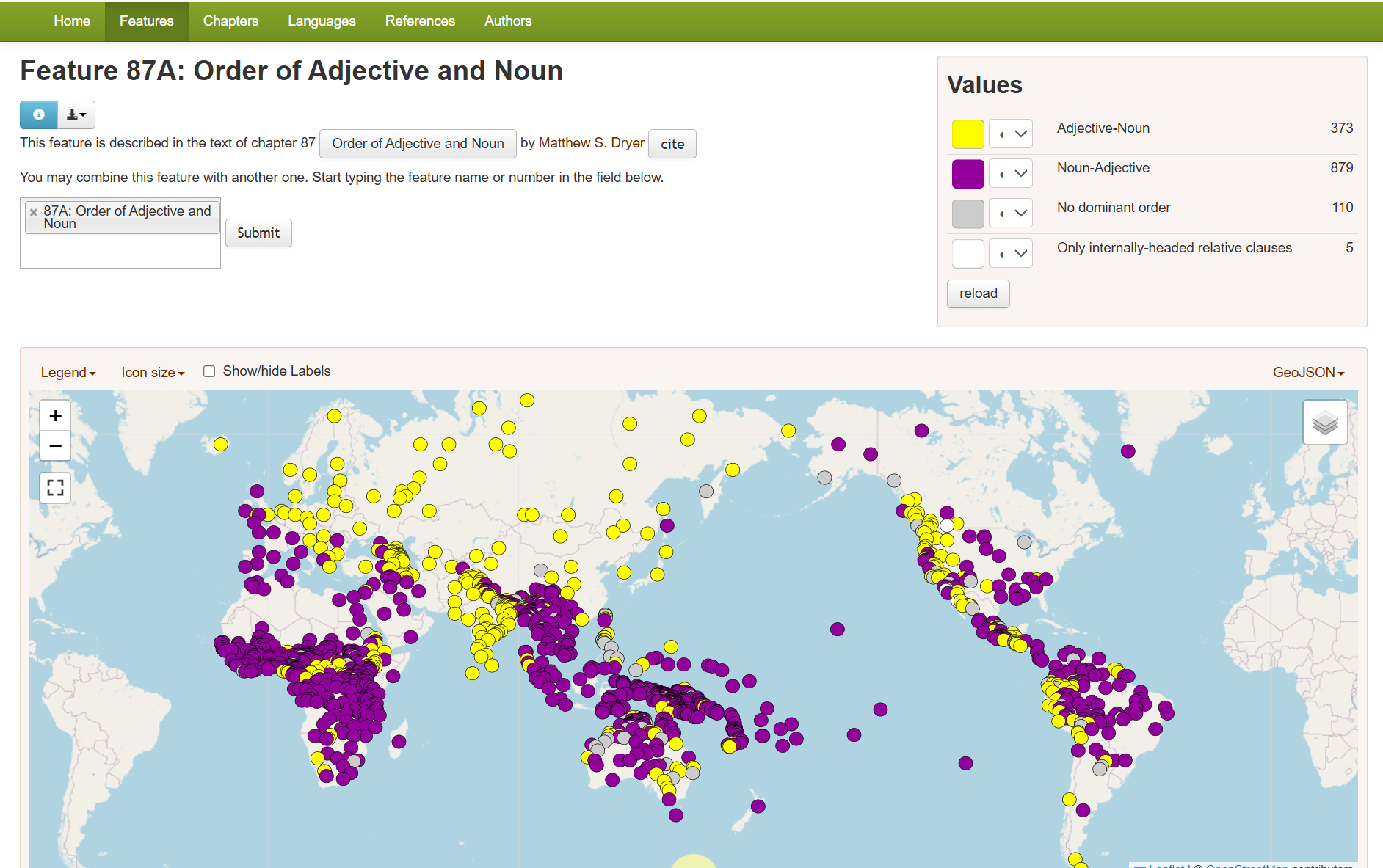}
\caption{A feature page from WALS Online illustrating how each feature defines a set of possible values (right panel) and maps their distribution across languages (bottom panel).}
\label{fig:wals-example}
\end{figure}

The World Atlas of Language Structures (WALS) is a large typological database documenting structural properties of the world's languages \citep{WALS}. It covers 192 features across 2,660 languages, with each language annotated for a subset of features based on available descriptive sources. Figure \ref{fig:wals-example} illustrates a typical WALS feature: for each feature, WALS defines a set of possible values and documents which value is attested in which languages.

We chose WALS as the basis for our benchmark because it provides human-verified ground-truth labels across a broad set of languages, including many low-resource ones, and its feature-value structure translates naturally into a multiple-choice QA format.  In the following subsections, we describe the language inventory and feature structure in more detail (Sections \ref{sec:wals-languages}-\ref{sec:wals-features}), then explain how we construct the benchmark (\Cref{sec:benchmark-construction}).

\subsection{Languages and samples in WALS}
\label{sec:wals-languages}

WALS contains data on 2,660 languages, each annotated with metadata such as genus, family, and ISO 639-3 code, enabling genealogical and geographical analyses. The languages are distributed across six major macro-areas: Africa, Eurasia, Papua and Oceania, North America, South America, and Australia. In addition to the full language inventory, the WALS authors also define a curated 100-language sample designed to maximize genealogical and areal diversity and mitigate biases arising from the over-representation of well-documented language families and regions. Because this sample exhibits substantially higher feature coverage (95–159 features per language), we use it as a complementary dataset in our language-level analyses (Section \ref{sec:performance-languages}) to disentangle effects of annotation sparsity from genuine cross-linguistic differences in model performance.Table~\ref{tab:wals-lang-statistics} summarizes the distribution of languages across macro-areas for both the full WALS database and the WALS 100-language sample.

\begin{table}[htbp]
\centering
\caption{Distribution of languages across macro-geographical areas in the full WALS database and the WALS 100-language sample.}
\label{tab:wals-lang-statistics}
\begin{tabular}{lccc}
\toprule
\textbf{Macroarea} & \textbf{WALS} & \textbf{WALS-100} \\
\midrule
Africa & $606$ & $17$ \\
Eurasia & $659$ & $28$ \\
Papua and Oceania & $560$ & $17$ \\
North America & $396$ & $18$ \\
South America & $258$ &  $13$ \\
Australia & $183$ & $7$ &  \\
\bottomrule
\end{tabular}
\end{table}

\subsection{Linguistic features and domains in WALS}
\label{sec:wals-features}

WALS documents 192 structural features that capture different aspects of grammatical organization across languages. Each feature is defined by a fixed set of discrete values representing alternative structural options. Languages are annotated with a single value per feature where data are available; for example, for \emph{Feature~33: Coding of Nominal Plurality}, the value attested for English is \emph{plural suffix} (see Figure \ref{fig:placeholder}), and for  \emph{Feature~87: Order of Adjective and Noun} it is \emph{adjective-noun} (Figure \ref{fig:wals-example}).

\begin{table}[htbp]
\centering
\caption{Distribution of WALS features across linguistic domains, showing the number of features per domain, the number of possible values per feature, and the number of languages for which each feature is attested (reported as minimum–maximum range with mean $\mu$).}

\setlength{\tabcolsep}{0pt}
\begin{tabular}{lclc}
\toprule
\textbf{Linguistic domain} & \textbf{Num. features} & \textbf{Num. values} & \textbf{Num. lang. per feat.} \\
\midrule
Word order & $56$ & $2$ - $28$ ($\mu = 7.61$) & $5$ - $1518$ \\
Nominal categories & $29$ & $2$ - $21$ ($\mu = 4.83$) & $71$ - $1066$ \\
Simple clauses & $26$ & $2$ - $23$ ($\mu = 5.92$) & $118$ - $1157$ \\
Phonology & $20$ & $2$ - $8$ ($\mu = 7.61$) & $40$ - $567$ \\
Verbal categories & $17$ & $2$ - $28$ ($\mu = 3.95$) & $193$ - $1131$ \\
Lexicon & $13$ & $2$ - $21$ ($\mu = 7.15$) & $72$ - $617$ \\
Morphology & $12$ & $2$ - $8$ ($\mu = 5.17$) & $145$ - $969$ \\
Nominal syntax & $8$ & $3$ - $8$ ($\mu = 6$) & $124$ - $301$ \\
Complex sentences & $7$ & $2$ - $7$ ($\mu = 4.86$) & $112$ - $283$ \\
Sign languages & $2$ & $3$ - $6$ ($\mu = 4.50$) & $35$ - $38$ \\
Clicks (Other) & $1$ & $4$ - $4$ ($\mu = 4$) & $143$ - $143$ \\
Writing systems (Other) & $1$ & $5$ - $5$ ($\mu = 5$) & $6$ - $6$ \\
\bottomrule
\end{tabular}
\label{tab:wals-feat-statistics}
\end{table}

The features are grouped into 12 linguistic domains, ranging from phonology and morphology to clause structure and word order (Table~\ref{tab:wals-feat-statistics}). Although the resulting language–feature matrix is sparse (i.e. not every feature is documented for every language), it contains as many as 76,475 manually verified data points. Because all annotations are curated by domain experts based on descriptive linguistic sources, WALS provides a reliable ground-truth resource for evaluating knowledge of specific structural properties across individual languages.

\subsection{Benchmark construction}
\label{sec:benchmark-construction}

To construct a multilingual benchmark from WALS, we transform its structured representation of linguistic features and annotations into a set of explicit metalinguistic benchmark items. Each WALS feature is mapped to a single grammatical question accompanied by a fixed set of answer options, and each item corresponds to a documented grammatical property of a specific language. The annotated WALS value for a given language–feature pair serves as the ground-truth label.

The resulting benchmark comprises 192 distinct question types, one for each WALS feature. These question types function as reusable templates and are instantiated across all languages for which WALS provides annotations, yielding a large set of language-specific question–answer pairs. Questions are derived from WALS feature descriptions, while answer options reflect the corresponding feature value categories. Below is an example for feature 129A (\textit{Hand and Arm}):

\vspace{1em}
\begin{adjustwidth}{2em}{2em}
\noindent\textbf{Question:} \emph{How are the concepts of 'hand' and 'arm' expressed in the X language?}

\vspace{0.5em}
\noindent\textbf{Answer options:}
\begin{itemize}[leftmargin=1.5em, itemsep=0.2em]
    \item \emph{Identity -- a single word denotes both 'hand' and 'arm'}
    \item \emph{Differentiation -- separate words denote 'hand' and 'arm'}
\end{itemize}
\end{adjustwidth}
\vspace{1em}

For more complex features, WALS encodes a large number of highly compressed and terminology-heavy value labels that combine multiple grammatical properties. For instance, Feature 144L (\textit{The Position of Negative Morphemes in SOV Languages}) distinguishes various patterns of negation placement using symbolic shorthand (e.g. \textit{NegSOV} as one of the possible values). To handle such cases, we systematically rephrased both feature names and value labels into clearer formulations that spell out the relevant grammatical configurations (e.g. \textit{What is the position of negative words in subject–object–verb clauses in the X language?} as the question, and \textit{Negative word before the subject, object, and verb} as one of its possible answers). This makes questions more interpretable for both models and readers, and provides some control over potential surface-level memorization effects—a point we return to in Section  \ref{sec:discussion}.

Each benchmark entry corresponds to a single linguistic feature and includes a feature identifier and name, a task question, a fixed set of possible answers, and a language-keyed map of ground-truth answers derived from WALS. Ground-truth annotations are provided only for languages attested for each feature. The dataset is split by feature rather than by language, with features stratified by linguistic domain and assigned to training, validation, and test splits to ensure balanced domain coverage and prevent feature leakage. The datasets are stored in JSON Lines (JSONL) format. The prompt is stored separately from the question content. The benchmark is released as an open-source dataset under the CC-BY-4.0 licence \footnote{A preliminary version is available at: \url{https://github.com/Oranzna/metalinguistic_benchmark}. The final version will be archived on CLARIN.SI and Hugging Face.}.

\section{Model setup and evaluation procedure}
\label{sec:setup}
This section first introduces the LLMs and prompting strategy used (\Cref{sec:llms-prompt}), and then presents the evaluation framework and metrics (\Cref{sec:eval-metrics}). We then describe a set of language-level properties used in our analyses (\Cref{sec:external-factors-descriptions}), such as measures of digital presence and language proximity, which are later examined as potential predictors of model performance.

\subsection{LLM models and prompting strategy}
\label{sec:llms-prompt}

We tested the benchmark on three models: one large proprietary (GPT-4o) and two large open-source models (Llama-3.3-70B; Gemma-3-27B). The GPT-4o model is chosen as a representative of current state-of-the-art models, while open models are evaluated to examine potential differences in behavior and support reproducibility of our experiments. Smaller models were not systematically evaluated, as they showed poor performance in initial tests, a finding that is consistent with prior research on linguistic knowledge in LLMs \citep{jumelet2025multiblimp}. 

We prompted GPT-4o through API requests, while the open-source models were run on a high-performance computing cluster for inference. We used a zero-shot prompting strategy, prompting the model with the feature question for each of the languages listed under each linguistic feature. For each WALS feature, models were prompted with a single feature-derived question instantiated for each language annotated for that feature, together with a fixed set of predefined answer options. This setup constrains the task to explicit selection among alternatives rather than free-form generation.

We set the temperature parameter to 0.2 to reduce randomness in model outputs and encourage consistent and deterministic behaviour across runs. This value was selected based on preliminary experiments. An example prompt is shown below:
\vspace{1em}
\begin{adjustwidth}{2em}{2em}
\emph{ “How large is the consonant inventory in the English language? The options are Small; Moderately small; Average; Moderately large; Large. Answer with one of the options only. Do not explain."}
\end{adjustwidth}

\subsection{Evaluation framework and metrics}
\label{sec:eval-metrics}

This section presents the evaluation metrics and how they are applied at three levels of analysis: overall model performance, performance across linguistic domains, and performance across individual languages.

\subsubsection{Overall performance evaluation}

We evaluated model performance using \textbf{accuracy} and \textbf{macro $F_1$} metrics. Accuracy was calculated separately for each feature as the proportion of correctly generated feature values across all languages for which that feature is documented. Since feature-value classes are frequently imbalanced, with some values occurring much more frequently than others, we also reported macro $F_1$ for each feature, which assigns equal weight to all classes and therefore provides a more balanced assessment of performance across both frequent and rare values.

To contextualize model performance, we compare accuracy against two simple baselines:

\begin{itemize}
    \item \textbf{Random chance baseline.} The expected accuracy from random selection among the available options. This varies by feature depending on the number of possible values (e.g., 50\% for binary features).
    \item \textbf{Majority class baseline.} The accuracy achieved by always predicting the most frequent value for a given feature. Performance above this baseline indicates that a model captures more than just the dominant pattern.
\end{itemize}

\subsubsection{Evaluation by linguistic domain}
To evaluate performance across linguistic domains, we computed weighted accuracy for each domain. Since features vary considerably in how many languages they cover, we weighted each feature's accuracy by the proportion of languages it represents within its domain. This means that broadly attested features—those documented across many languages—contribute more to domain-level scores, providing more robust estimates of model performance than features with sparse coverage.

To enable fair comparison across domains with different baseline difficulties, we also computed relative accuracy gain over the majority-class %and chance
baseline at the feature level, then aggregated using the same weighting procedure:

\begin{equation}
\text{\textbf{Relative Accuracy Gain}}(f) = \frac{\text{Accuracy}_f - \text{Baseline}_f}{\text{Baseline}_f}
\end{equation}

\noindent where \textit{f} denotes a feature. This normalises for the fact that some domains have higher majority-class baselines than others (i.e. are inherently easier to predict due to more skewed value distributions).

\subsubsection{Evaluation by language}

Language performance was measured as the proportion of linguistic features that a model answered correctly out of the total number of features present for that language in WALS.

Direct comparison of model performance at the level of individual languages is challenging due to highly uneven feature coverage in WALS: many languages are annotated for only a small number of features, making per-language accuracy estimates unstable and difficult to interpret. Consequently, ranking all languages would conflate model performance with annotation sparsity. To address this, we adopted a two-stage approach.

First, we perform a coarse-grained analysis by grouping languages according to digital status following the six-class taxonomy of \citet{joshi-etal-2020-state}, which categorises languages based on the availability of labelled and unlabelled resources, ranging from class 0 (very low digital presence, no unlabelled data) to class 5 (dominant digital presence, significant resource investment). This allows us to assess how metalinguistic performance varies with digital support at the group level, aggregating accuracy within each status category rather than comparing individual languages. We perform this analysis on both the full WALS dataset and the WALS 100-language sample.

Second, for the WALS 100-language sample, which provides substantially denser and more uniform annotation (95–159 features per language), we additionally report the top- and bottom-performing languages per model as illustrative examples of language-level variation.

\subsection{Identifying external factors associated with model performance}
\label{sec:external-factors-descriptions}

To investigate which factors are associated with variation in model performance, we examine external variables at two levels of analysis. At the domain level (Section \ref{sec:methodology-search-engine-frequency}), we analyse the online visibility of individual linguistic features. At the language level (Section \ref{sec:methodology-lang-predictors}, we examine a set of language-level predictors spanning linguistic, sociolinguistic, and resource-related dimensions.

\subsubsection{Domain-level predictor}
\label{sec:methodology-search-engine-frequency}

We check whether model performance across different linguistic domains is related to the online footprint of each of the 192 WALS features. The WALS feature name serves as the search keyword; if the name is too general (e.g. \textit{tone}), it is refined to ensure linguistic relevance (e.g. \textit{tone in language}). Although Google search engine result counts are approximations, they serve as a reasonable proxy for the presence of linguistic features in online texts. We obtain approximate hit counts using the Google Search API, which provides reproducible result estimates.

We compute the Pearson correlation coefficient ($r$) between domain-level accuracy and the average number of hits for features within each domain. We apply a $\log_{10}$ transformation to search result counts to account for their wide range and to enable meaningful comparison across domains with very different levels of online prevalence.

\subsubsection{Language-level predictors}
\label{sec:methodology-lang-predictors}

To examine which factors are associated with language-level performance, we consider eight predictors spanning various dimensions related to digital presence, sociolinguistic status, and linguistic relatedness. Several of these predictors capture partially overlapping aspects of language use and visibility; accordingly, our analysis focuses on their relative importance rather than treating them as independent causal factors. To ensure comparability across languages, we restrict this analysis to the WALS 100-language sample. We consider the following language-level predictors:

\begin{itemize}

    \item \textbf{Resource availability.}  
    We use the aforementioned digital status taxonomy proposed by \citet{joshi-etal-2020-state}, which classifies languages into six categories based on the availability of labelled and unlabelled resources, ranging from very low digital presence (e.g. Bora) to dominant digital presence (e.g. Spanish)

    \item \textbf{Digital language support.}  
    Ethnologue’s global digital language support scale\footnote{\url{https://www.ethnologue.com/}} classifies languages into five levels, from \textit{still} (no digital support) to \textit{thriving} (supported by advanced tools, including AI).

    \item \textbf{Language vitality.}  
    Ethnologue’s vitality scale classifies languages into four levels based on intergenerational transmission and institutional use, ranging from \textit{institutional} (the language is used in institutions outside of home and community) to \textit{extinct} (the language is no longer used).

    \item \textbf{Wikipedia size.}  
    Wikipedia\footnote{https://wikistats.wmcloud.org/display.php?t=wp} size is used as an indicator of a language’s digital presence. Languages with more articles are usually better represented in digital environments and have a more active digital community. We choose the number of articles as an indicator of digital language presence.

    \item \textbf{UD corpus size.}  
    The size of a language’s Universal Dependencies (UD) treebanks \citep{de-marneffe-etal-2021-universal} indicates the availability of curated, grammatically annotated resources and reflects the degree of attention the language has received in computational linguistics research. We use the number of tokens available per language in the latest UD release \citep{ud215}.

    \item \textbf{Geographical macroregion.}  
    The broad geographic area where a language is primarily spoken, following the six macroareas defined in WALS (Africa, Eurasia, Papunesia, Australia, North America, South America).

    \item \textbf{Language family.}  
    We use the top-level genealogical family assignments provided by WALS, which cover all languages in the 100-language sample. WALS distinguishes major language families (e.g. Indo-European, Niger–Congo, Austronesian), treating language isolates as single-language families. For example, English is classified as Indo-European.

    \item \textbf{Proximity to English.} We include typological distance measures based on lang2vec representations \citep{littell2017uriel}, following the distance-based analysis framework of \citet{van2025distals}. Lang2vec encodes languages as vectors of typological features, allowing us to quantify structural similarity to English and assess whether such similarity is associated with higher model performance. 
    
\end{itemize}

To assess the relative importance of these predictors, we divide languages into three accuracy groups (high, middle, low) and train a random forest classifier to predict group membership using 10-fold cross-validation. Cross-validated performance was assessed with the Matthews correlation coefficient (MCC). Random forests provide interpretable feature-importance scores, allowing us to determine which language-level factors are most strongly associated with model performance. For selected predictors, we additionally report Spearman’s ($\rho$) to quantify monotonic relationships with accuracy.

\section{Results and analysis}
\label{sec:results}
In this section, we present the results of evaluating three LLMs using the newly constructed WALS-based benchmark (Section \ref{sec:wals}) and experimental setup (Section \ref{sec:setup}). We begin by examining overall performance (\Cref{sec:overall-performance}), then analyse performance across linguistic domains and its relationship with online feature visibility (\Cref{sec:performance-ling-areas}), and finally examine performance across languages and which factors predict language-level variation (\Cref{sec:performance-languages}).

\subsection{Overall LLM performance}
\label{sec:overall-performance}

We first examined LLM performance across individual features by computing accuracy and macro $F_1$ for each feature. Overall model performance is reported as the unweighted mean of these feature scores.

Overall performance is low across all models (Table~\ref{tab:llm-performance}). GPT-4o achieved the highest accuracy (0.367), followed by Llama-3.3-70B (0.265) and Gemma-3-27B (0.246), with the same ranking for macro $F_1$. All models perform well above the chance baseline (0.234), indicating that they capture some systematic regularities rather than guessing at random. However, none outperform the majority-class baseline (0.539), meaning their predictions fail to improve upon simply selecting the most frequent feature value.

Together, these results show that metalinguistic question answering remains a challenging task for current LLMs. While models capture broad grammatical regularities, their knowledge reflects dominant cross-linguistic patterns rather than fine-grained, language-specific distinctions.

\begin{table}[htbp]
\centering
\caption{Overall LLM performance on the WALS-based metalinguistic benchmark. We report unweighted mean accuracy and macro $F_1$ across all 192 grammatical features.}
\label{tab:llm-performance}
\begin{tabular}{lcccc}
\toprule
\textbf{LLM model} & \textbf{Accuracy} & \textbf{Macro $F_1$} \\
\midrule
Chance baseline & $0.234$ \\
Majority-class baseline & $0.539$  \\[5pt]
GPT-4o          & $0.367$ & $0.228$ \\
Llama-3.3-70B  & $0.265$ & $0.157$ \\
Gemma-3-27B    & $0.246$ & $0.129$ \\
\bottomrule
\end{tabular}

\end{table}

\subsection{LLM performance across linguistic domains}
\label{sec:performance-ling-areas}

\begin{figure}[htbp]
\centering
\includegraphics[width=1\linewidth]{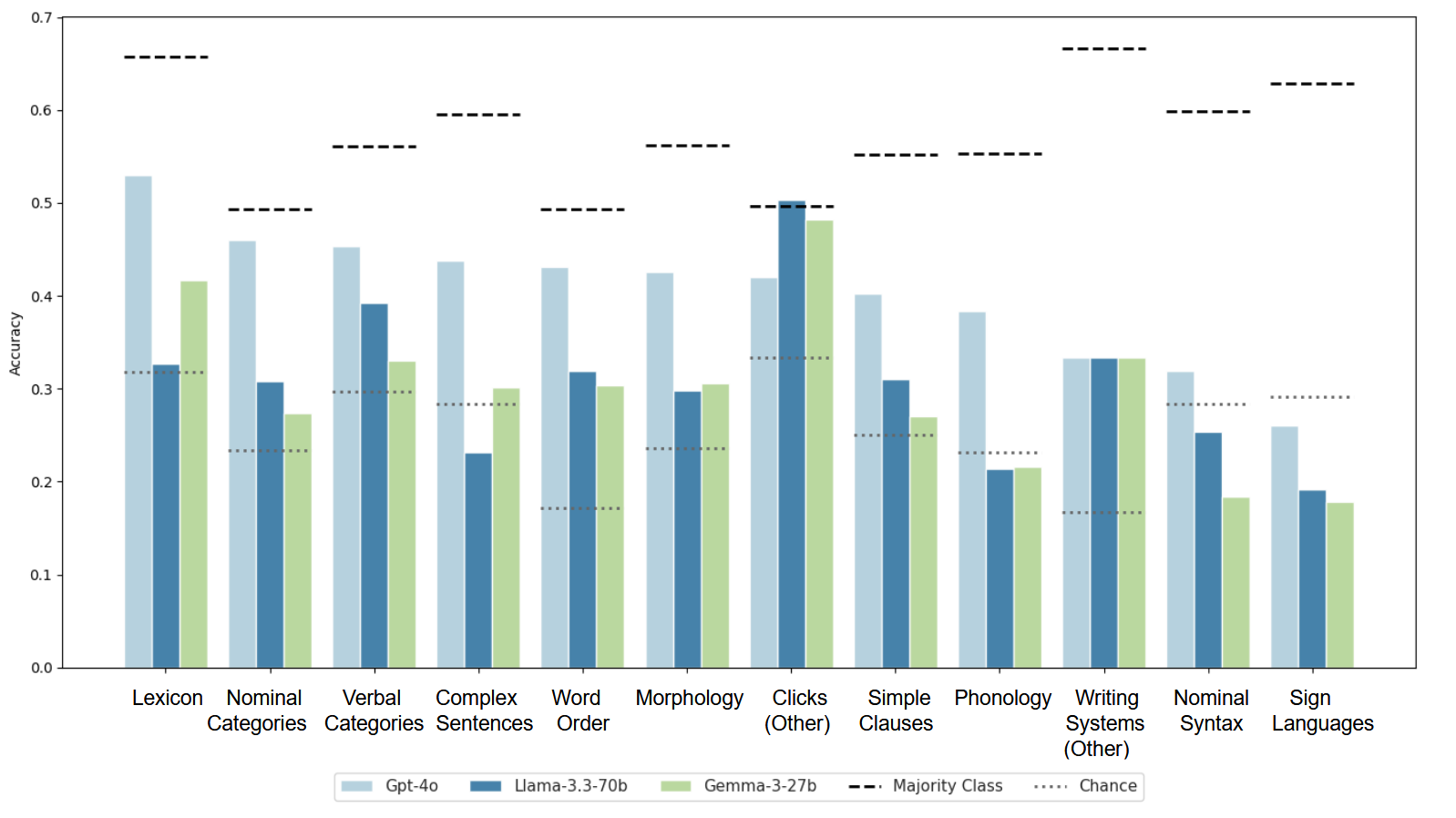}
\vspace{10pt}
\caption{Normalized LLM accuracy across linguistic domains relative to majority-class and chance baselines, ranked by GPT-4o performance.}
\label{fig:norm-acc-ling-areas-all}
\end{figure}

We next examined how model performance varies across linguistic domains (\Cref{fig:norm-acc-ling-areas-all}). Accuracy was highest for questions related to lexicon and verbal categories, and lowest for phonology, nominal syntax, and sign languages. This pattern was consistent across all three models, though GPT-4o additionally showed moderately strong performance on nominal categories.

Relative accuracy gains over the majority-class baseline - which account for differing baseline difficulties across domains - confirm this pattern. All domains show negative gains, but the magnitude varies substantially. For GPT-4o, the smallest deficits appear for nominal categories (-0.03) and morphology (-0.14), while the largest deficits appear for sign languages (-0.59) and nominal syntax (-0.47). The pattern is similar for the other models, with sign languages, nominal syntax, and phonology consistently showing the weakest performance across all three LLMs.

\begin{figure}[htbp]
\centering
\includegraphics[width=0.9\linewidth]{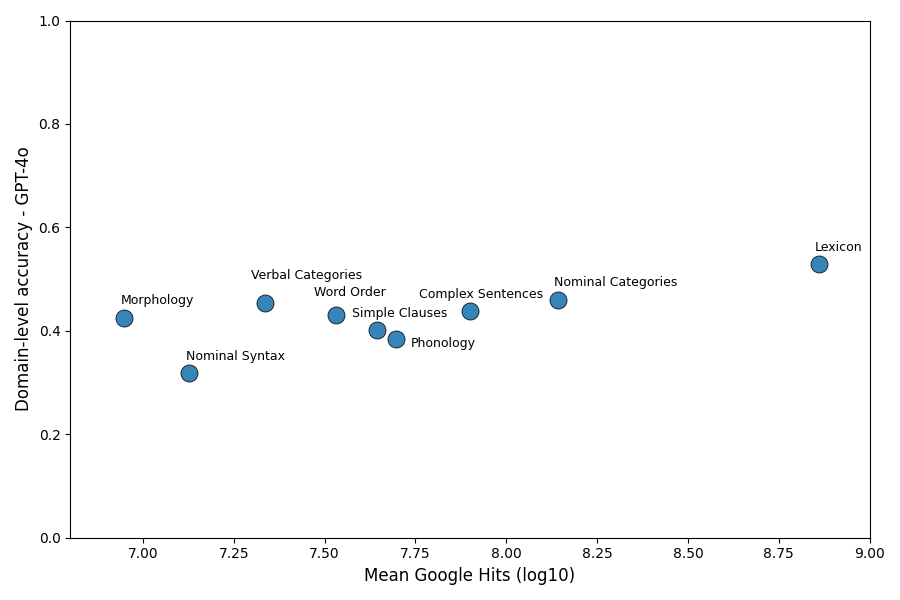}
\vspace{10pt}
\caption{Correlation between domain-level accuracy and online visibility (mean Google hits per domain) for GPT-4o (r = 0.715).}
\label{fig:corr_acc_domain_gpt}
\end{figure}

To investigate whether this variation reflects differences in how well linguistic phenomena are represented online, we computed the correlation between domain-level accuracy and mean Google search hit counts for features within each domain (see \Cref{sec:methodology-search-engine-frequency}). We excluded three domains containing only one or two features (Clicks, Writing Systems, Sign Languages), as their estimates are less stable. For GPT-4o, accuracy correlates strongly with online visibility (r = 0.715; \Cref{fig:corr_acc_domain_gpt}), with a similar pattern for Gemma-3-27B (r = 0.571). No such relationship was observed for Llama-3.3-70B (r = 0.045),  which may be attributed to differences in training data composition and curation strategies.

Together, these results show that metalinguistic knowledge in LLMs is unevenly distributed across linguistic domains, with some domains substantially easier than others. Our analysis suggests that the online visibility of linguistic phenomena may partially account for this variation, although the relationship is not consistent across models and warrants further investigation.

\subsection{LLM performance across languages}
\label{sec:performance-languages}

Finally, we examine how LLM performance varies across languages, focusing on performance by digital language status (\Cref{sec:status-results}), illustrative top- and bottom-performing languages (\Cref{sec:lang-top-bottom}), and language-level predictors of model accuracy (\Cref{sec:lang-predictors}).

\subsubsection{Performance by language status}
\label{sec:status-results}

\begin{figure*}[htbp]
\centering
\begin{subfigure}[t]{0.48\textwidth}
  \centering
  \includegraphics[width=\linewidth]{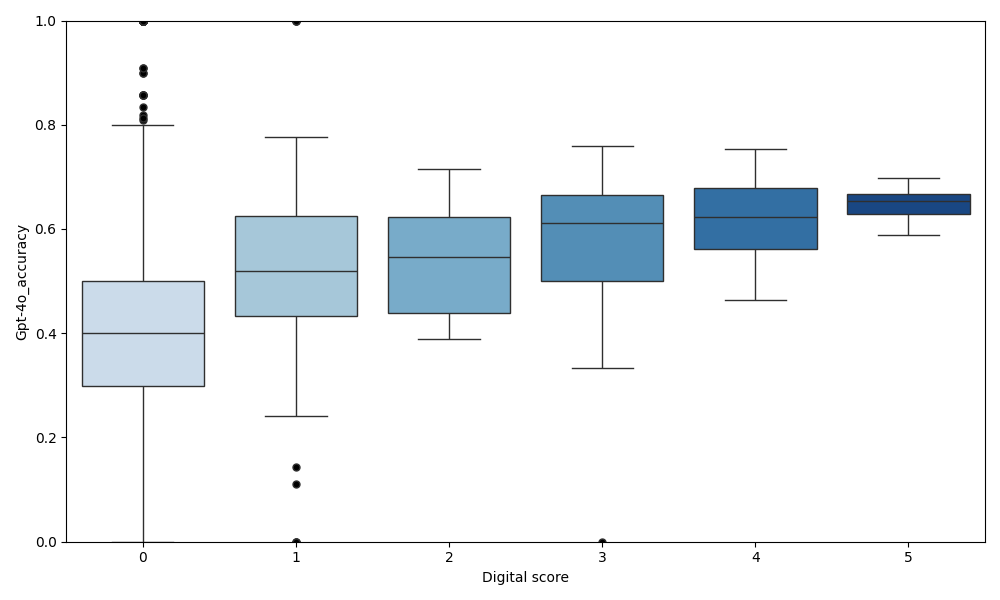}
  \caption{All WALS languages}
  \label{fig:corr_gpt}
\end{subfigure}\hfill
\begin{subfigure}[t]{0.48\textwidth}
  \centering
  \includegraphics[width=\linewidth]{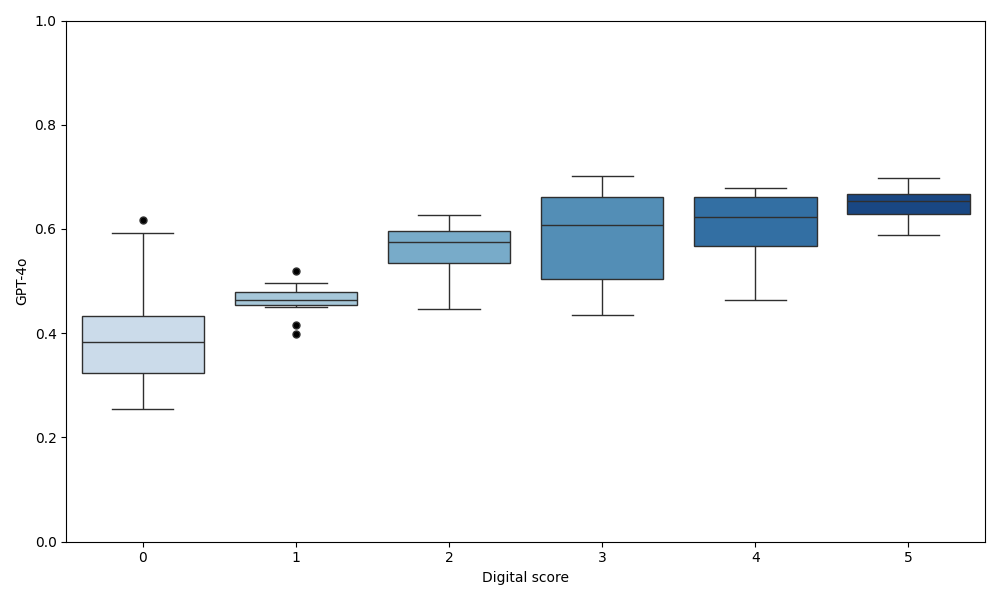}
  \caption{WALS 100-language sample}
  \label{fig:corr_gpt_100wals}
\end{subfigure}
\caption{Distribution of accuracy by digital status (0 = very low to 5 = dominant) for GPT-4o.}
\label{fig:corr_gpt_all}
\end{figure*}

\begin{figure*}[htbp]
\centering
\begin{subfigure}[t]{0.48\textwidth}
  \centering
  \includegraphics[width=\linewidth]{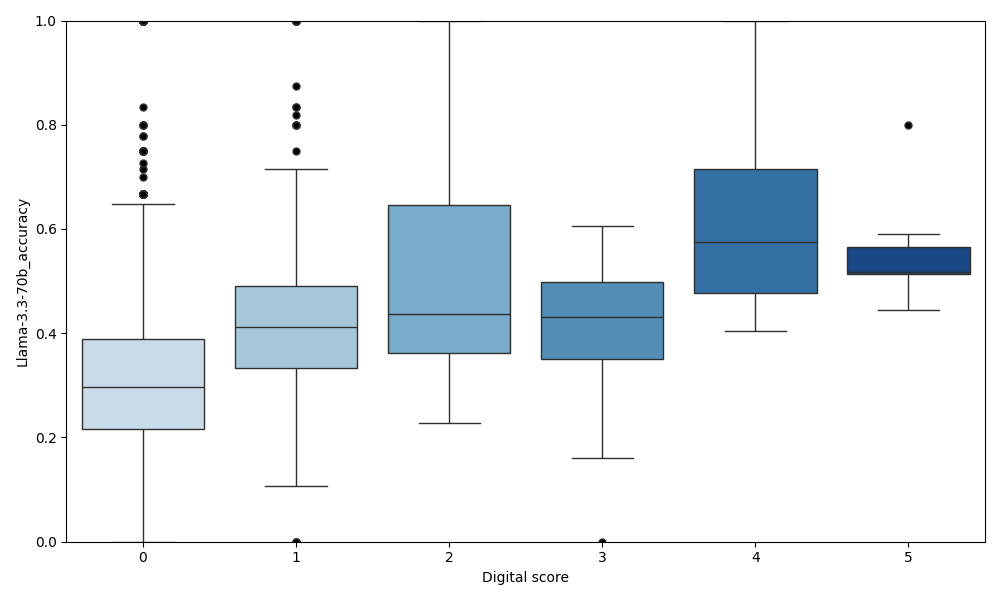}
  \caption{All WALS languages}
  \label{fig:corr_llama}
\end{subfigure}\hfill
\begin{subfigure}[t]{0.48\textwidth}
  \centering
  \includegraphics[width=\linewidth]{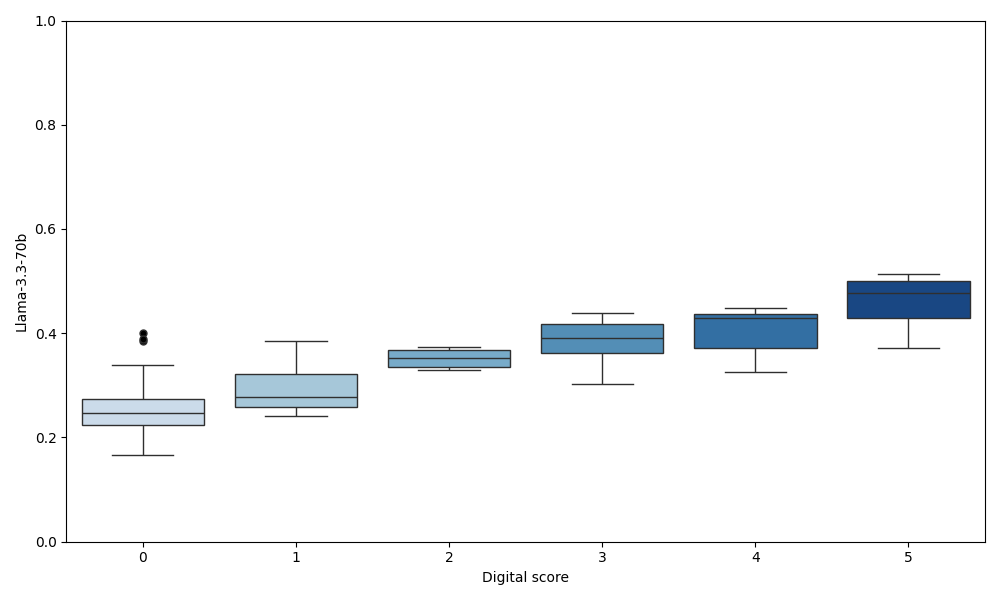}
  \caption{WALS 100-language sample}
  \label{fig:corr_llama_100wals}
\end{subfigure}
\caption{Distribution of accuracy by digital status (0 = very low to 5 = dominant) for Llama-3.3-70b.}
\label{fig:corr_lamma_all}
\end{figure*}

\begin{figure*}[htbp]
\centering
\begin{subfigure}[t]{0.48\textwidth}
  \centering
  \includegraphics[width=\linewidth]{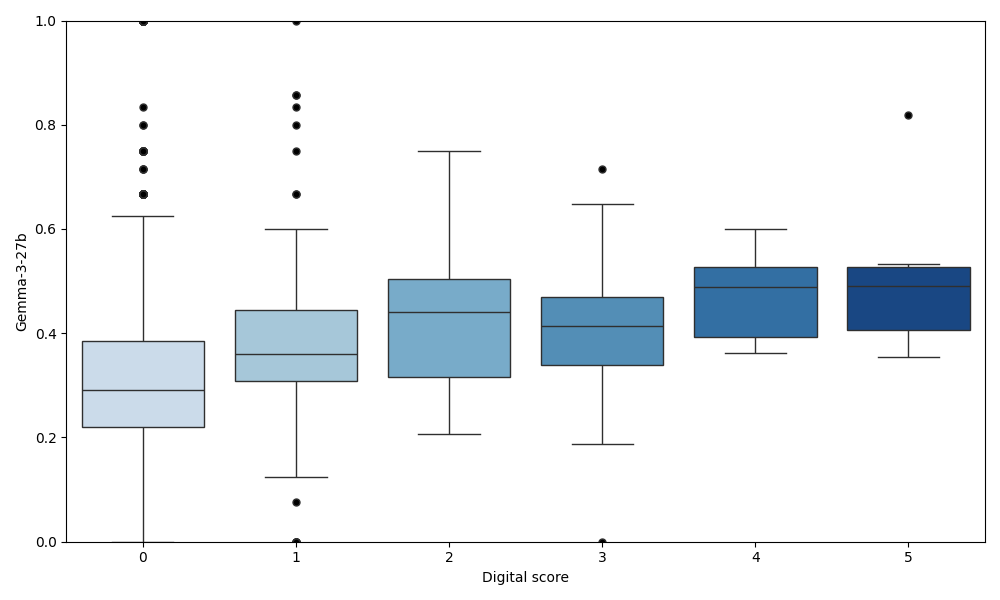}
  \caption{All WALS languages}
  \label{fig:corr_gemma}
\end{subfigure}\hfill
\begin{subfigure}[t]{0.48\textwidth}
  \centering
  \includegraphics[width=\linewidth]{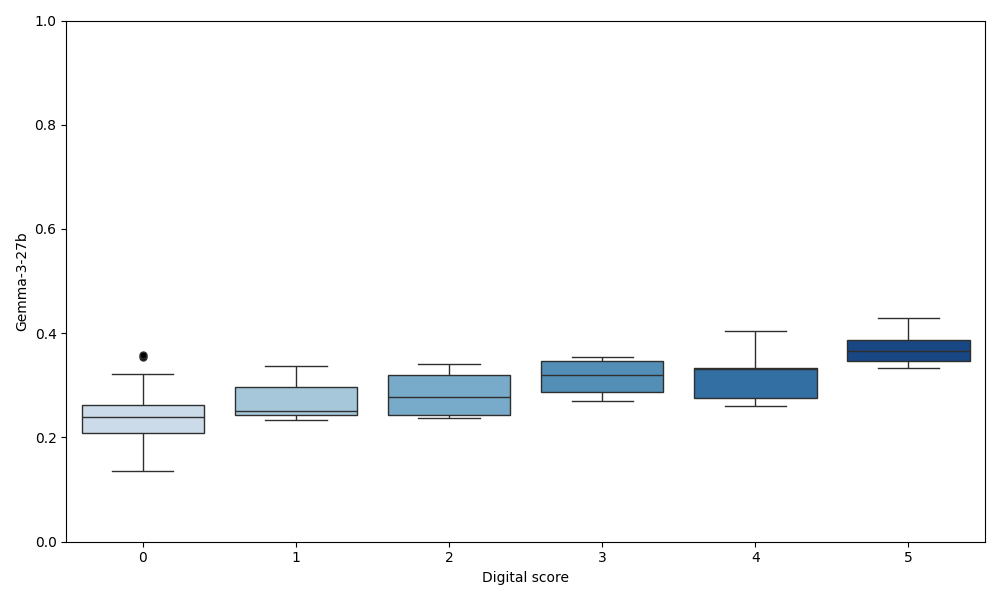}
  \caption{WALS 100-language sample}
  \label{fig:corr_gemma_100wals_gemma}
\end{subfigure}
\caption{Distribution of accuracy by digital status (0 = very low to 5 = dominant) for Gemma-3-27b.}
\label{fig:corr_gemma_all}
\end{figure*}

To examine the relationship between model performance and language resourcedness, we grouped languages by digital status according to the six-class taxonomy of \citet{joshi-etal-2020-state}: 0 = very low digital presence, no unlabelled data (2,191 languages, e.g., Bora); 1 = low, some unlabelled data (222 languages, e.g., Navajo); 2 = low, some labelled data (19 languages, e.g., Zulu); 3 = moderate, insufficient labelled data (28 languages, e.g., Hebrew); 4 = strong, large unlabelled but less labelled data (18 languages, e.g., Hungarian); 5 = dominant, significant resource investment (7 languages, e.g., Spanish).
Figures~\ref{fig:corr_gpt_all}–\ref{fig:corr_gemma_all} present the distribution of mean accuracy by digital status for all three models, shown for both the full WALS dataset and the 100-language sample. A clear pattern emerges: languages with higher digital status achieve higher accuracy across all models. Because digital status is ordinal, we report Spearman's correlation ($\rho$). Correlations are relatively weak for the full dataset (GPT-4o: $\rho = 0.227$; Llama-3.3-70B: $\rho = 0.23$; Gemma-3-27B: $\rho = 0.182$), but substantially stronger for the 100-language sample (GPT-4o: $\rho = 0.734$, Llama-3.3-70B: $\rho = 0.710$ and Gemma-3-27B: $\rho = 0.598$).

These results reveal three key patterns. First, metalinguistic performance varies substantially across languages: models consistently struggle more with some languages than others. Second, resource availability emerges as a strong predictor of this variation across all three models — languages with limited digital presence perform worse regardless of model architecture or size. Third, GPT-4o, the largest model in our comparison, achieves the highest accuracy overall, with the advantage most pronounced for well-resourced languages. This suggests that increased model capacity amplifies the benefit of abundant training data, but does not compensate for the lack of it.

\subsubsection{Top- and bottom-performing languages}
\label{sec:lang-top-bottom}

To complement the population-level analysis, we next examine performance differences between individual languages. Since direct per-language comparison is unreliable for the full WALS dataset—where some languages are annotated for only a handful of features—we restrict this analysis to the WALS 100-language sample, which provides a more uniform feature coverage. 

Tables \ref{tab:top_ten} and \ref{tab:bottom_ten} list the ten highest- and lowest-performing languages for each model. Across all models, the top-performing languages are predominantly high-resource, digitally well-supported languages such as English, German, French, Spanish, and Mandarin. In contrast, the bottom-performing languages are largely low-resource languages with limited digital presence, including Barasano, Imonda, Wichí, and Kutenai — a pattern consistent with the status analysis above (Figures~\ref{fig:corr_gpt_all}–\ref{fig:corr_gemma_all}).

\setlength{\tabcolsep}{3pt}
\begin{table}[htbp]
\centering
\caption{Top ten languages by accuracy for each model (WALS 100-language sample).}
\label{tab:top_ten}
\begin{tabular}{l r l r l r}
\toprule
\textbf{GPT-4o} & \textbf{Accuracy} & \textbf{Llama-3.3-70b} & \textbf{Accuracy} & \textbf{Gemma-3-27b} & \textbf{Accuracy} \\
\midrule
Hebrew (M) & $0.702$ & 
French & $0.513$ & Spanish & $0.429$ \\
English & $0.698$ & 
English & $0.503$ & Turkish & $0.405$ \\
Russian & $0.680$ & 
Mandarin & $0.497$ & English & $0.392$ \\
German & $0.675$ & 
German & $0.478$ & French & $0.382$ \\
Thai & $0.669$ & 
Vietnamese & $0.448$ & German & $0.365$ \\
Finnish & $0.665$ & 
Spanish & $0.445$ & Korean & $0.358$ \\
Spanish & $0.658$ & 
Turkish & $0.442$ & Indonesian & $0.355$ \\
Vietnamese & $0.657$ & 
Hebrew (M) & $0.440$ & Mandarin & $0.354$ \\
Mandarin & $0.654$ & 
Finnish & $0.432$ & Kannada & $0.354$ \\
French & $0.653$ & 
Russian & $0.430$ & Greek (M) & $0.351$ \\
\bottomrule
\end{tabular}
\end{table}

\setlength{\tabcolsep}{3pt}
\begin{table}[htbp]
\centering
\caption{Bottom ten languages by accuracy for each model (WALS 100-language sample).}
\label{tab:bottom_ten}
\begin{tabular}{l r l r l r}
\toprule
\textbf{GPT-4o} & \textbf{Accuracy} & \textbf{Llama-3.3-70b} & \textbf{Accuracy} & \textbf{Gemma-3-27b} & \textbf{Accuracy} \\
\midrule
Barasano & $0.313$ & 
Canela & $0.199$ & Mixtec & $0.176$ \\
Lavukaleve & $0.311$ & 
Otomí & $0.193$ & Maybrat & $0.173$ \\
Rama & $0.307$ & 
Apurin\~{a} & $0.192$ & Maricopa & $0.165$ \\
Nama & $0.307$ & 
Otomí & $0.193$ & Karok & $0.162$ \\
Alamblak & $0.306$ & 
Paiwan & $0.190$ & Canela & $0.162$ \\
Wari' & $0.306$ & 
Kutenai & $0.188$ & Imonda & $0.160$ \\
Wichí & $0.302$ & 
Wichí & $0.186$ & Slave & $0.155$ \\
Sanuma & $0.298$ & 
Mixtec & $0.182$ & Apurin\~{a} & $0.154$ \\
Imonda & $0.278$ & 
Maybrat & $0.179$ & Lakhota & $0.149$ \\
Maricopa & $0.254$ & 
Kayardild & $0.167$ & Kutenai & $0.136$ \\
\bottomrule
\end{tabular}
\end{table}

\subsubsection{Predictors of language-level performance}
\label{sec:lang-predictors}

\begin{figure}[htbp]
\centering
\includegraphics[width=\linewidth]{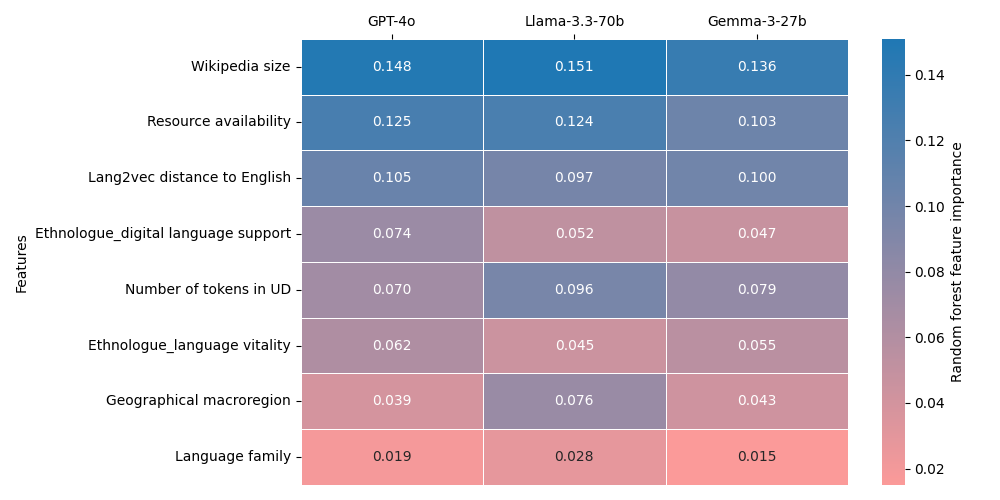}
\vspace{10pt}
\caption{Random forest feature importance for predicting language-level accuracy group (high, middle, low) across three models.}
\label{fig:features_rank}
\end{figure}

To examine which factors best predict language-level accuracy, we trained a random forest classifier to predict performance group (high, middle, low) based on the eight predictors described in \Cref{sec:methodology-lang-predictors}.\footnote{We measure association, which may not necessarily translate into causation.} Model training was performed using 10-fold cross validation. Cross-validated performance, evaluated using the MCC, was 0.581 for GPT-4o, 0.589 for Llama-3.3-70b, and 0.403 for Gemma-3-27b, indicating a moderate-to-strong association between the predictors and performance group. Figure~\ref{fig:features_rank} shows the feature importance rankings for all three models.

Resource-related factors emerge as the strongest predictors across all models. Wikipedia size ranks highest for all three (GPT-4o: 0.148; Llama-3.3-70B: 0.151; Gemma-3-27B: 0.136), followed by resource availability (GPT-4o: 0.125; Llama-3.3-70B: 0.124; Gemma-3-27B: 0.103). This is consistent with the digital-status analysis above: languages with larger digital footprints are easier for models to answer questions about, likely because more descriptive and metalinguistic content about these languages is available in training data.

Notably, Lang2vec distance to English ranks third across all models (GPT-4o: 0.105; Llama-3.3-70B: 0.097; Gemma-3-27B: 0.100), ahead of sociolinguistic factors such as language vitality and geographical macroregion. This suggests that typological similarity to English — the dominant language in LLM training data — provides an additional advantage beyond mere resource availability. We explain this by noting that English is likely the main source of training data for the tested LLMs.

In contrast, language family is consistently the weakest predictor (GPT-4o: 0.019; Llama-3.3-70B: 0.028; Gemma-3-27B: 0.015), indicating that genealogical relatedness contributes little to model accuracy when resource-related factors are taken into account. Sociolinguistic factors such as Ethnologue language vitality and geographical macroregion fall in the middle range, suggesting that while these factors play some role, they are less informative than direct measures of digital presence.

Together, these results indicate that LLM metalinguistic performance is shaped primarily by data availability, with typological proximity to English as a secondary factor. Genealogical and sociolinguistic properties of languages, by contrast, explain relatively little of the variation.

\section{Discussion}
\label{sec:discussion}

In this work, we introduced a massively multilingual benchmark for evaluating metalinguistic knowledge in LLMs, derived from grammatical features documented in the World Atlas of Language Structures. Using this benchmark, we evaluated three contemporary LLMs across nearly two hundred linguistic features and more than 2,600 languages. Our results confirm earlier findings from smaller-scale studies that LLMs exhibit limited explicit grammatical knowledge but extend them to a global scale and a much broader range of linguistic domains. We show that metalinguistic performance varies substantially across domains and languages, with particularly low accuracy for phonological and syntactically complex features, and systematic disadvantages for low-resource languages across all models. Beyond these findings, the benchmark opens many opportunities for further exploration leveraging the rich human-curated knowledge encoded in WALS: analyses targeting specific domains, geographical regions, or language families; correlations with external factors beyond those examined here; and systematic comparison across a wider range of models and prompting strategies.

At the same time, using WALS as a benchmarking resource entails several methodological limitations. As a typological database, WALS was designed to capture broad structural distinctions relevant for cross-linguistic comparison, rather than to provide exhaustive grammatical descriptions. Its feature inventory reflects particular descriptive traditions and theoretical choices, which could be expanded or complemented in future benchmarks. Second, feature coverage is sparse and uneven, making generalizations difficult; we address this by using the WALS 100-language sample, though future benchmarks could target feature subsets with more uniform attestation across languages (or draw on similar resources like Grambank \citep{Grambank2023}, which offers more systematic per-language coverage, though without the phonological and lexical domains examined here). Third, WALS represents grammatical properties as discrete feature values, which do not necessarily reflect the gradient, context-dependent patterns of language use observed in corpora \citep{YanLiu2023,Levshina2023,Baylor2023}. While discrete categories provide a clear evaluation target, future benchmarks could complement them with corpus-based representations that capture gradience and variation in actual language use \citep[e.g.,][]{klemen2025towards, Baylor2024}.

The use of WALS also introduces some experimental considerations. Because WALS is publicly available, models may have encountered its content during training. While low overall accuracy and systematic variation across domains and languages suggest that direct retrieval is not a significant factor, future work could introduce additional controls such as paraphrased questions and answers across the full set of features (see Section \ref{sec:benchmark-construction}) or newly added features. A further limitation is the multiple-choice prompt format. Recent evidence suggests that models may exploit option-level artifacts, such as elimination heuristics or surface cues in the answer choices, leading them to select correct options without fully solving the underlying task \cite{raman2025reasoning}. However, the low accuracy observed here suggests that such prompt-induced effects were not primary drivers of model performance.

Despite these limitations, our benchmark provides an unprecedented new resource for evaluating LLMs across low-resource and underdocumented languages. Our findings reveal that limited digital presence affects not only performance on standard NLP tasks, but also models' explicit knowledge about language itself—a critical gap given the growing interest in using LLMs to support language documentation and preservation. By making such disparities visible at a global scale, our work underscores the importance of inclusive evaluation frameworks and demonstrates the value of leveraging human-curated linguistic knowledge to probe model behaviour beyond surface language use. We release the benchmark as an open-source dataset to support this goal.

\section{Conclusion}
\label{sec:conclusion}

We introduced a massively multilingual benchmark for evaluating metalinguistic knowledge in LLMs, drawing on the linguistic features and languages documented in WALS.  Our evaluation of three contemporary models reveals that metalinguistic knowledge in current LLMs is limited and fragmentary: accuracy is low overall, varies systematically across linguistic domains, and correlates strongly with resource availability. These findings suggest that current LLMs mainly reflect the distribution of digitally available data rather than exhibiting generalizable grammatical knowledge across the world's languages.

By scaling metalinguistic evaluation to over 2,600 languages, this study provides the most comprehensive assessment of LLMs' explicit linguistic knowledge to date, revealing that the languages most in need of computational support are precisely those about which models know the least. We release the benchmark as an open-source resource to support further research on metalinguistic evaluation across the world's languages.

\section*{Acknowledgments}
The work was primarily supported by the Large Language Models for Digital Humanities project (GC-0002), funded by the Slovene Research and Innovation Agency (ARIS), and the ARIS core research programme P6-0411. Additional support was provided by the EU ERA Chair grant no. 101186647 (AI4DH).

\vspace{0.5em}
\noindent Competing interests: None.

\vspace{0.5em}
\noindent Generative AI tools were used to assist with language editing.

\bibliographystyle{nlplike}
\bibliography{WALSbib}

@article{klemen2025towards,
  title={Towards corpus-grounded agentic {LLMs} for multilingual grammatical analysis},
  author={Klemen, Matej and Ar{\v{c}}on, Tja{\v{s}}a and Ter{\v{c}}on, Luka and Robnik-{\v{S}}ikonja, Marko and Dobrovoljc, Kaja},
  journal={arXiv preprint arXiv:2512.00214},
  year={2025}
}

@article{YanLiu2023,
  author  = {Jianwei Yan and Haitao Liu},
  title   = {Basic Word Order Typology Revisited: A Crosslinguistic Quantitative Study Based on {UD} and {WALS}},
  journal = {Linguistics Vanguard},
  volume  = {9},
  number  = {1},
  pages   = {73--85},
  year    = {2023},
  doi     = {10.1515/lingvan-2021-0001},
  url     = {https://doi.org/10.1515/lingvan-2021-0001}
}

@article{Levshina2023,
  author  = {Natalia Levshina and Savithry Namboodiripad and Marc Allassonnière-Tang and Mathew Kramer and Luigi Talamo and Annemarie Verkerk and Sasha Wilmoth and Gabriela Garrido Rodriguez and Timothy Michael Gupton and Evan Kidd and Zoey Liu and Chiara Naccarato and Rachel Nordlinger and Anastasia Panova and Natalia Stoynova},
  title   = {Why we need a gradient approach to word order},
  journal = {Linguistics},
  volume  = {61},
  number  = {4},
  pages   = {825--883},
  year    = {2023},
  doi     = {10.1515/ling-2021-0098},
  url     = {https://doi.org/10.1515/ling-2021-0098}
}

@article{de-marneffe-etal-2021-universal,
    title = "{U}niversal {D}ependencies",
    author = "de Marneffe, Marie-Catherine  and
      Manning, Christopher D.  and
      Nivre, Joakim  and
      Zeman, Daniel",
    journal = "Computational Linguistics",
    volume = "47",
    number = "2",
    month = jun,
    year = "2021",
    address = "Cambridge, MA",
    publisher = "MIT Press",
    url = "https://aclanthology.org/2021.cl-2.11/",
    doi = "10.1162/coli_a_00402",
    pages = "255--308"
}

@inproceedings{singh-etal-2023-explaining,
    title = "Explaining data patterns in natural language with language models",
    author = "Singh, Chandan  and
      Morris, John X.  and
      Aneja, Jyoti  and
      Rush, Alexander  and
      Gao, Jianfeng",
    editor = "Belinkov, Yonatan  and
      Hao, Sophie  and
      Jumelet, Jaap  and
      Kim, Najoung  and
      McCarthy, Arya  and
      Mohebbi, Hosein",
    booktitle = "Proceedings of the 6th BlackboxNLP Workshop: Analyzing and Interpreting Neural Networks for NLP",
    xmonth = dec,
    year = "2023",
    xaddress = "Singapore",
    xpublisher = "Association for Computational Linguistics",
    url = "https://aclanthology.org/2023.blackboxnlp-1.3/",
    doi = "10.18653/v1/2023.blackboxnlp-1.3",
    pages = "31--55"
}

@article{annurev:/content/journals/10.1146/annurev-linguistics-031220-120504,
   author = "Berez-Kroeker, Andrea L. and Gabber, Shirley and Slayton, Aliya",
   title = "Recent advances in technologies for resource creation and mobilization in language documentation", 
   journal= "Annual Review of Linguistics",
   year = "2023",
   volume = "9",
   number = "Volume 9, 2023",
   pages = "195-214",
   doi = "https://doi.org/10.1146/annurev-linguistics-031220-120504",
   url = "https://www.annualreviews.org/content/journals/10.1146/annurev-linguistics-031220-120504",
   publisher = "Annual Reviews",
   issn = "2333-9691",
   type = "Journal Article",
  }

@misc{tanzer2024benchmarklearningtranslatenew,
      title={A benchmark for learning to translate a new language from one grammar book}, 
      author={Garrett Tanzer and Mirac Suzgun and Eline Visser and Dan Jurafsky and Luke Melas-Kyriazi},
      year={2024},
      eprint={2309.16575},
      archivePrefix={arXiv},
      primaryClass={cs.CL},
      url={https://arxiv.org/abs/2309.16575}, 
}

@inproceedings{joshi-etal-2020-state,
    title = "The state and fate of linguistic diversity and inclusion in the {NLP} world",
    author = "Joshi, Pratik  and
      Santy, Sebastin  and
      Budhiraja, Amar  and
      Bali, Kalika  and
      Choudhury, Monojit",
    editor = "Jurafsky, Dan  and
      Chai, Joyce  and
      Schluter, Natalie  and
      Tetreault, Joel",
    booktitle = "Proceedings of the 58th Annual Meeting of the Association for Computational Linguistics",
    xmonth = jul,
    year = "2020",
    xaddress = "Online",
    xpublisher = "Association for Computational Linguistics",
    url = "https://aclanthology.org/2020.acl-main.560/",
    doi = "10.18653/v1/2020.acl-main.560",
    pages = "6282--6293"
}

@article{begus2025large,
  title={Large linguistic models: investigating {LLMs’} metalinguistic abilities},
  author={Beguš, Gašper and Dabkowski, Maksymilian and Rhodes, Ryan},
  journal={IEEE Transactions on Artificial Intelligence},
  year={2025},
  publisher={IEEE}
}

@misc{ud215,
  author = {Daniel Zeman and Joakim Nivre and Mitchell Abrams and Elia Ackermann and No{\'e}mi Aepli and Hamid Aghaei and {\v Z}eljko Agi{\'c} and Amir Ahmadi and Lars Ahrenberg and ...},
  title  = {Universal Dependencies 2.15},
  year   = {2024},
  url    = {http://hdl.handle.net/11234/1-5787},
  note   = {{LINDAT}/{CLARIAH}-{CZ} digital library at the Institute of Formal and Applied Linguistics ({{\'U}FAL}), Faculty of Mathematics and Physics, Charles University}
}

@article{Grambank2023,
  author  = {Hedvig Skirgård and Hannah J. Haynie and Damián E. Blasi and Harald Hammarström and Jeremy Collins and Jay J. Latarche and Jakob Lesage and Tobias Weber and Alena Witzlack-Makarevich and Sam Passmore and Angela Chira and Luke Maurits and Russell Dinnage and Michael Dunn and Ger Reesink and Ruth Singer and Claire Bowern and Patience Epps and Jane Hill and Outi Vesakoski and Martine Robbeets and Noor Karolin Abbas and Daniel Auer and Nancy A. Bakker and Giulia Barbos and Robert D. Borges and Swintha Danielsen and Luise Dorenbusch and Ella Dorn and John Elliott and Giada Falcone and Jana Fischer and Yustinus Ghanggo Ate and Hannah Gibson and Hans-Philipp Göbel and Jemima A. Goodall and Victoria Gruner and Andrew Harvey and Rebekah Hayes and Leonard Heer and Roberto E. Herrera Miranda and Nataliia Hübler and Biu Huntington-Rainey and Jessica K. Ivani and Marilen Johns and Erika Just and Eri Kashima and Carolina Kipf and Janina V. Klingenberg and Nikita König and Aikaterina Koti and Richard G. A. Kowalik and Olga Krasnoukhova and Nora L.M. Lindvall and Mandy Lorenzen and Hannah Lutzenberger and Tônia R.A. Martins and Celia Mata German and Suzanne van der Meer and Jaime Montoya Samamé and Michael Müller and Saliha Muradoglu and Kelsey Neely and Johanna Nickel and Miina Norvik and Cheryl Akinyi Oluoch and Jesse Peacock and India O.C. Pearey and Naomi Peck and Stephanie Petit and Sören Pieper and Mariana Poblete and Daniel Prestipino and Linda Raabe and Amna Raja and Janis Reimringer and Sydney C. Rey and Julia Rizaew and Eloisa Ruppert and Kim K. Salmon and Jill Sammet and Rhiannon Schembri and Lars Schlabbach and Frederick W.P. Schmidt and Amalia Skilton and Wikaliler Daniel Smith and Hilário de Sousa and Kristin Sverredal and Daniel Valle and Javier Vera and Judith Voß and Tim Witte and Henry Wu and Jingting Ye and Maisie Yong and Tessa Yuditha and Roberto Zariquiey and Robert Forkel and Nicholas Evans and Stephen C. Levinson and Martin Haspelmath and Simon J. Greenhill and Quentin D. Atkinson and Russell D. Gray},
  title   = {Grambank reveals the importance of genealogical constraints on linguistic diversity and highlights the impact of language loss},
  journal = {Science Advances},
  volume  = {9},
  number  = {16},
  year    = {2023},
  doi     = {10.1126/sciadv.adg6175}
}

@inproceedings{Baylor2024,
  author    = {Emi Baylor and Esther Ploeger and Johannes Bjerva},
  title     = {Multilingual gradient word-order typology from {U}niversal {D}ependencies},
  booktitle = {Proceedings of the 18th Conference of the European Chapter of the Association for Computational Linguistics (Volume 2: Short Papers)},
  editor    = {Yvette Graham and Matthew Purver},
  xaddress   = {St. Julian's, Malta},
  xpublisher = {Association for Computational Linguistics},
  year      = {2024},
  month     = mar,
  pages     = {42--49},
  url       = {https://aclanthology.org/2024.eacl-short.6/}
}

@article{kellert2025parsing,
  title={Parsing the switch: {LLM-based UD} Annotation for complex code-switched and low-resource languages},
  author={Kellert, Olga and Tyagi, Nemika and Imran, Muhammad and Licona-Guevara, Nelvin and G{\'o}mez-Rodr{\'\i}guez, Carlos},
  journal={arXiv preprint arXiv:2506.07274},
  year={2025}
}

@inproceedings{ramji2025inductive,
  title={Inductive linguistic reasoning with large language models},
  author={Ramji, Raghav and Ramji, Keshav},
  booktitle={Findings of the Association for Computational Linguistics: ACL 2025},
  pages={22783--22810},
  year={2025}
}

@inproceedings{ide2025make,
  title={How to make the most of {LLMs’} grammatical knowledge for acceptability judgments},
  author={Ide, Yusuke and Nishida, Yuto and Vasselli, Justin and Oba, Miyu and Sakai, Yusuke and Kamigaito, Hidetaka and Watanabe, Taro},
  booktitle={Proceedings of the 2025 Conference of the Nations of the Americas Chapter of the Association for Computational Linguistics: Human Language Technologies (Volume 1: Long Papers)},
  pages={7416--7432},
  year={2025}
}

@article{jumelet2025multiblimp,
  title={{MultiBLiMP 1.0: A} massively multilingual benchmark of linguistic minimal pairs},
  author={Jumelet, Jaap and Weissweiler, Leonie and Nivre, Joakim and Bisazza, Arianna},
  journal={arXiv preprint arXiv:2504.02768},
  year={2025}
}

@article{waldis2024holmes,
  title="{Holmes: A} benchmark to assess the linguistic competence of language models",
  author={Waldis, Andreas and Perlitz, Yotam and Choshen, Leshem and Hou, Yufang and Gurevych, Iryna},
  journal={Transactions of the Association for Computational Linguistics},
  volume={12},
  pages={1616--1647},
  year={2024},
  xpublisher={MIT Press 255 Main Street, 9th Floor, Cambridge, Massachusetts 02142, USA~…}
}

@inproceedings{Baylor2023,
  author    = {Emi Baylor and Esther Ploeger and Johannes Bjerva},
  title     = {The Past, Present, and Future of Typological Databases in {NLP}},
  booktitle = {Findings of the Association for Computational Linguistics: EMNLP 2023},
  editor    = {Houda Bouamor and Juan Pino and Kalika Bali},
  address   = {Singapore},
  publisher = {Association for Computational Linguistics},
  year      = {2023},
  month     = dec,
  pages     = {1163--1169},
  doi       = {10.18653/v1/2023.findings-emnlp.82},
  url       = {https://aclanthology.org/2023.findings-emnlp.82/}
}

@inproceedings{zhang2024mela,
  title={{MELA: Multilingual} evaluation of linguistic acceptability},
  author={Zhang, Ziyin and Liu, Yikang and Huang, Weifang and Mao, Junyu and Wang, Rui and Hu, Hai},
  booktitle={Proceedings of the 62nd Annual Meeting of the Association for Computational Linguistics (Volume 1: Long Papers)},
  pages={2658--2674},
  year={2024}
}

@inproceedings{behzad2023elqa,
  title={{ELQA: A} corpus of metalinguistic questions and answers about {E}nglish},
  author={Behzad, Shabnam and Sakaguchi, Keisuke and Schneider, Nathan and Zeldes, Amir},
  booktitle={Proceedings of the 61st Annual Meeting of the Association for Computational Linguistics (Volume 1: Long Papers)},
  pages={2031--2047},
  year={2023}
}

@inproceedings{spencer-kongborrirak-2025-llms,
  author    = {Spencer, Piyapath T. and Kongborrirak, Nanthipat},
  title     = {Can {LLMs} Help Create Grammar?: {Automating} Grammar Creation for Endangered Languages with In-Context Learning},
  booktitle = {Proceedings of the 31st International Conference on Computational Linguistics},
  pages     = {10214--10227},
  year      = {2025},
  month     = jan,
  xaddress   = {Abu Dhabi, UAE},
  xpublisher = {Association for Computational Linguistics},
  url       = {https://aclanthology.org/2025.coling-main.681/}
}

@misc{WALS,
  author    = {Matthew S. Dryer and Martin Haspelmath},
  publisher = {Zenodo},
  title     = {{WALS} Online (v2020.4)},
  type      = {Data set},
  url       = {https://doi.org/10.5281/zenodo.13950591},
  year      = {2013},
  doi       = {10.5281/zenodo.13950591}
}

@article{lu2024ai,
  title={The {AI Scientist: Towards} fully automated open-ended scientific discovery},
  author={Lu, Chris and Lu, Cong and Lange, Robert Tjarko and Foerster, Jakob and Clune, Jeff and Ha, David},
  journal={arXiv preprint arXiv:2408.06292},
  year={2024}
}

@article{Chang2024SurveyLLM,
  author  = {Chang, Yupeng and Wang, Xu and Wang, Jindong and Wu, Yuan and
             Yang, Linyi and Zhu, Kaijie and Chen, Hao and Yi, Xiaoyuan and
             Wang, Cunxiang and Wang, Yidong and others},
  title   = {A Survey on Evaluation of Large Language Models},
  journal = {ACM Transactions on Intelligent Systems and Technology},
  volume  = {15},
  number  = {3},
  articleno = {39},
  pages   = {1--45},
  year    = {2024},
  month   = mar,
  doi     = {10.1145/3641289},
  url     = {https://doi.org/10.1145/3641289}
}

@inproceedings{suvarna2404phonologybench,
    title = "{P}honology{B}ench: Evaluating Phonological Skills of Large Language Models",
    author = "Suvarna, Ashima  and
      Khandelwal, Harshita  and
      Peng, Nanyun",
    booktitle = "Proceedings of the 1st Workshop on Towards Knowledgeable Language Models (KnowLLM 2024)",
    month = aug,
    year = "2024",
    doi = "10.18653/v1/2024.knowllm-1.1",
    pages = "1--14"
}

@inproceedings{yang2025linggym,
  title={{LingGym: How far are LLMs} from thinking like field linguists?},
  author={Yang, Changbing and Ma, Franklin and Shi, Freda and Zhu, Jian},
  booktitle={Proceedings of the 2025 Conference on Empirical Methods in Natural Language Processing},
  pages={1314--1340},
  year={2025}
}

@article{goyal2025iolbench,
  title={{IOLBENCH: Benchmarking LLMs} on Linguistic Reasoning},
  author={Goyal, Satyam and Dan, Soham},
  journal={arXiv preprint arXiv:2501.04249},
  year={2025}
}

@article{thrush2024strange,
  title={I am a strange dataset: Metalinguistic tests for language models},
  author={Thrush, Tristan and Moore, Jared and Monares, Miguel and Potts, Christopher and Kiela, Douwe},
  journal={arXiv preprint arXiv:2401.05300},
  year={2024}
}

@article{wang2025cpg,
  title={{CPG-EVAL: A} Multi-Tiered Benchmark for Evaluating the {Chinese} Pedagogical Grammar Competence of Large Language Models},
  author={Wang, Dong},
  journal={arXiv preprint arXiv:2504.13261},
  year={2025}
}

@inproceedings{xu2025can,
  title={Can large language models be good language teachers?},
  author={Xu, LiQing and Li, Qiwei and Peng, Tianshuo and Li, Zuchao and Zhao, Hai and Wang, Ping},
  booktitle={Proceedings of the 2025 Conference on Empirical Methods in Natural Language Processing},
  pages={23968--23982},
  year={2025}
}

@inproceedings{van2025distals,
  title={{DistaLs: A} Comprehensive Collection of Language Distance Measures},
  author={Van Der Goot, Rob and Ploeger, Esther and Blaschke, Verena and Samardzic, Tanja},
  booktitle={Proceedings of the 2025 Conference on Empirical Methods in Natural Language Processing: System Demonstrations},
  pages={307--318},
  year={2025}
}

@article{lian2025lingbench++,
  title={{LingBench++: A} linguistically-informed benchmark and reasoning framework for multi-step and cross-cultural inference with {LLMs}},
  author={Lian, Da-Chen and Huang, Ri-Sheng and Chen, Pin-Er and Lim, Chunki and Lin, You-Kuan and Tseng, Guan-Yu and Yang, Zi-Cheng and Lin, Zhen-Yu and Chen, Pin-Cheng and Hsieh, Shu-Kai},
  journal={arXiv preprint arXiv:2507.16809},
  year={2025}
}

@inproceedings{littell2017uriel,
  title     = {{URIEL and lang2vec: Representing }languages as typological, geographical, and phylogenetic vectors},
  author    = {Littell, Patrick and Mortensen, David R. and Lin, Ke and Kairis, Katherine and Turner, Carlisle and Levin, Lori},
  booktitle = {Proceedings of the 15th Conference of the European Chapter of the Association for Computational Linguistics: Volume 2, Short Papers},
  year      = {2017},
  pages     = {8--14},
  address   = {Valencia, Spain},
  publisher = {Association for Computational Linguistics}
}

@article{raman2025reasoning,
  title={Reasoning models are test exploiters: Rethinking multiple-choice},
  author={Raman, Narun and Lundy, Taylor and Leyton-Brown, Kevin},
  journal={arXiv preprint arXiv:2507.15337},
  year={2025}
}

\label{lastpage}

\end{document}